\documentclass[11pt,a4paper]{article}

\usepackage[T1]{fontenc}
\usepackage[utf8]{inputenc}
\usepackage{lmodern}
\usepackage[margin=1in]{geometry}

\usepackage{amsmath,amssymb}
\usepackage{graphicx}
\usepackage{booktabs}
\usepackage{tabularx}
\usepackage{siunitx}
\usepackage{xcolor}
\usepackage{caption}
\usepackage{subcaption}
\usepackage{float}
\usepackage{placeins}

\usepackage{natbib}
\usepackage{hyperref}
\usepackage[nameinlink,capitalize]{cleveref}

\title{SepsisAI Orchestrator: A Containerized and Scalable Platform for Deploying AI Models and Real-Time Monitoring in Early Sepsis Detection}

\author{
Santiago Ospitia\textsuperscript{1} \and
John Sanabria\textsuperscript{1} \and
John García-Henao\textsuperscript{2,3}
\\[0.5em]
\textsuperscript{1}School of Systems Engineering and Computing, University of Valle, Cali, Colombia\\
\textsuperscript{2}Digital Medicine Unit, Balgrist University Hospital, Zurich, Switzerland\\
\textsuperscript{3}Nucleus-AI Research, Medellin, Colombia
}

\date{}

\begin{document}

\maketitle

\begin{abstract}
Despite strong predictive results in the clinical machine learning literature, the translation of these models into bedside use remains limited
by systems-level barriers: heterogeneous data representations, the absence of standardized deployment workflows, and a mismatch between research prototypes and the concurrency and latency requirements of hospital environments. We present the SepsisAI-Orchestrator, an open-source modular platform that addresses this deployment gap for early sepsis detection. The platform integrates HL7 FHIR-inspired Clinical Document Architecture (CDA) preprocessing, NoSQL storage, a containerized LightGBM classifier served via REST APIs, and a Streamlit clinical dashboard, orchestrated with Docker and Kubernetes. A previously validated LightGBM model (F1 0.87-0.94 on PhysioNet 2019) is reused without modification; the contribution lies in the surrounding infrastructure and its empirical characterization under load. Using k6 with 50-1000 concurrent virtual users, we find that replica count must be matched to the physical CPU thread count of the host: scaling from 3 to 12 replicas on a 12-thread CPU reduces p95 latency from 3.3\,s to 1.41\,s (57.3\,\% reduction) and eliminates all request failures, while over-provisioning to 24 or 48 replicas degrades performance due to scheduler contention. To our knowledge this U-shaped scaling behavior has not been quantified previously for clinical AI inference workloads. We do not claim prospective clinical validation. Source code and
deployment manifests are available at \url{https://github.com/nucleusai/sepsisai-orchestrator}.
\end{abstract}

\section{\label{sec:introduction}Introduction}

Despite a decade of strong predictive results in the clinical machine learning literature, the gap between published models and bedside deployment remains wide, and the bottleneck is increasingly recognized as systems-level rather than algorithmic~\cite{mitchell2025rebooting, esteva2019}. Heterogeneous data representations across hospitals, the absence of standardized deployment workflows, and mismatches between research prototypes and the concurrency, latency, and fault-tolerance requirements of hospital environments jointly
prevent otherwise accurate models from delivering clinical value. Sepsis, a life-threatening condition that affected more than 50 million people worldwide and caused approximately 11 million deaths in 2017~\cite{brant2022developing}, is a paradigmatic instance of this gap: it is high-stakes, time-critical, requires continuous monitoring, and depends on heterogeneous data sources, thereby exercising every component of a realistic clinical AI deployment pipeline.

This work addresses the deployment gap as a systems-for-machine-learning problem, with early sepsis detection as the concrete case study. General-purpose model-serving stacks such as KServe, Seldon Core, BentoML, and NVIDIA Triton, as well as healthcare-oriented stacks such as MONAI Deploy and Clara, provide important building blocks. What remains missing is an end-to-end, openly available reference architecture that ties these pieces together for clinical AI specifically, combining HL7 FHIR-based interoperability, container orchestration, real-time inference, and clinician-facing visualization and that empirically characterizes how such a system behaves under realistic hospital concurrency. We position our contribution at this gap; a detailed review of sepsis prediction models and clinical AI deployment infrastructure is deferred to Section~\ref{sec:state_of_art}.

We present the \textbf{SepsisAI-Orchestrator}, an open-source, modular platform that standardizes raw electronic health records into HL7 FHIR-inspired Clinical Document Architecture (CDA) structures, stores them in a NoSQL database, exposes a containerized LightGBM classifier via REST APIs, and provides a Streamlit-based clinical dashboard. Docker is used for portability and Kubernetes for orchestration, with each component mapped to an appropriate resource type (Deployments for stateless compute, StatefulSet for the database, Jobs for batch preprocessing). The LightGBM classifier is taken from prior work~\cite{toro2022mlfhir}; the contribution lies in the deployment infrastructure around it and in its empirical characterization under load. Evaluation on the PhysioNet/Computing in Cardiology Challenge 2019 dataset~\cite{reyna2020physionet} with k6 load tests of 50-1000 concurrent virtual users yields our central finding: replica count must be matched to the physical CPU thread count of the host. Scaling from 3 to 12 replicas on a 12-thread CPU reduced p95 latency from 3.3\,s to 1.41\,s (57.3\% reduction) and eliminated all request failures, whereas over-provisioning to 24 or 48 replicas degraded performance due to context-switching and scheduler contention. While CPU oversubscription is well known in systems literature, to our knowledge the resulting U-shaped scaling curve has not been quantified for clinical AI
inference workloads on commodity hardware.

The contributions of this work are:
\begin{itemize}
    \item An end-to-end open-source reference architecture for clinical AI deployment, integrating HL7 FHIR-inspired preprocessing, NoSQL storage, containerized REST-served inference, and a clinician-facing dashboard, released publicly at \url{https://github.com/nucleusai/sepsisai-orchestrator}.
    \item An empirical characterization of horizontal scaling for containerized clinical ML inference under realistic concurrency (50-1000 virtual users), revealing a U-shaped latency curve and identifying CPU-thread count as the practical optimum for replica provisioning on single-node deployments.
    \item A reproducible deployment recipe (Docker images, Kubernetes manifests, load-testing scripts) that practitioners can adapt to clinical prediction tasks beyond sepsis.
\end{itemize}

We intentionally scope the contribution to deployment systems rather than predictive modeling, and we do not claim prospective clinical validation. The platform is designed in alignment with HIPAA and GDPR principles through container isolation, role separation, and controlled data access; clinical deployment is left to future work. The remainder of the paper is organized as follows. Section~\ref{sec:state_of_art} reviews related work on sepsis prediction models and clinical AI deployment infrastructure. Section~\ref{sec:materials_methods} describes the architecture and services of the SepsisAI-Orchestrator. Section~\ref{sec:experiments} presents the experimental setup and the scalability analysis. Section~\ref{sec:conclusions} discusses implications, limitations, and future work.
\section{State of the Art}
\label{sec:state_of_art}
The prediction of sepsis with clinical data is an active research area that has progressed quickly in recent years. The literature can be grouped into two main perspectives: predictive models and infrastructures that support their deployment and monitoring in real environments. This section reviews both perspectives and identifies the main research gaps.

Initial methods for sepsis risk assessment used scoring systems such as SOFA and qSOFA, but their sensitivity and specificity were limited \cite{fahrmeir2021}. The availability of large scale EHR datasets made it possible to apply ML models such as logistic regression, Support Vector Machines (SVM), Random Forest, and XGBoost, which performed better than traditional scores \cite{henry2015}. More recent studies applied deep learning techniques including Recurrent Neural Networks (RNN), Long Short Term Memory (LSTM), and Gated Recurrent Units (GRU). Komorowski et al. \cite{komorowski2018} used reinforcement learning to predict sepsis and suggest treatment strategies. Henry et al. \cite{henry2015} showed that time series models can better forecast septic shock compared to static predictors. However, the lack of transparency of deep models has limited clinical trust, and explainable AI (XAI) methods such as those proposed by Tonekaboni et al. \cite{tonekaboni2019} are needed to improve adoption.

Strong predictive models cannot be used in practice without infrastructures that allow secure, efficient, and reliable deployment. This challenge is often described as the ``lab to live gap'' \cite{kent2020}, and the following are some of the most
relevant issues related to this gap: 
Health information systems use heterogeneous formats and standards, which create silos and errors. Standards such as HL7 FHIR, SNOMED CT, and OMOP CDM were developed to unify health data \cite{brat2020, torab2023, nan2023}. Effective interoperability, however, requires infrastructures that can handle diverse data flows \cite{yoo2022}.
MLOps applies DevOps principles to ML workflows. It provides reproducibility, traceability, and compliance with regulatory requirements such as HIPAA and GDPR. MLOps in healthcare must also monitor model drift, ensure data quality, and support reliable deployment of AI models \cite{kent2020}.
Docker allows applications to run with all their dependencies, while Kubernetes provides orchestration, scaling, and fault tolerance. These technologies enable high availability, workload balancing, and portability across cloud and on premise infrastructures. In clinical environments they make it possible to run patient specific predictions in parallel and ensure continuity of service under failures \cite{dias2019, nan2023}.

The literature shows agreement in three areas: (1) deep learning dominates predictive modeling for sepsis, (2) HL7 FHIR is the most accepted interoperability standard, and (3) container based MLOps is the foundation for clinical AI deployment. 
Still, most studies treat these aspects separately. 
Few works provide an integrated infrastructure that connects data preprocessing, interoperability, predictive models, and clinical dashboards. 
This research addresses this gap by presenting a modular and scalable infrastructure that supports sustainable deployment of AI for early sepsis detection.


\section{\label{sec:materials_methods}Materials and Methods}

\subsection{Study Design}
This study followed an experimental design in which we integrated a previously developed ML-based model for sepsis prediction, using missing data imputation and FHIR interoperability \cite{toro2022mlfhir}, into the modular platform of the \textbf{SepsisAI-Orchestrator}. The platform addresses interoperability, scalability, and adoption challenges by combining containerization with Docker and orchestration with Kubernetes. It implements the full workflow for early sepsis detection, including data preprocessing, model deployment, and real-time monitoring. The design consisted of four stages: (i) data preprocessing and standardization, (ii) AI model deployment, (iii) orchestration with Kubernetes, and (iv) clinical monitoring through a dashboard.

\subsection{Dataset}
We used the dataset released in the PhysioNet/Computing in Cardiology Challenge 2019 \cite{reyna2020physionet}, which contains time series clinical records from intensive care patients in three hospital systems. The dataset includes hourly measurements of vital signs (8 attributes), laboratory tests (26 attributes), and demographic variables (7 attributes). Data from two hospitals were made publicly available for training and validation, while a third was reserved for hidden testing. The dataset is imbalanced, with fewer septic than non septic patient records, and contains missing values.

\subsection{SepsisAI-Orchestrator: A Containerized and Scalable Platform}

To address the interoperability, scalability, and adoption challenges
identified in Section~\ref{sec:state_of_art}, we propose the
\textbf{SepsisAI-Orchestrator}, a modular infrastructure for AI deployment
and real-time monitoring in early sepsis detection. The platform was designed
in two versions. The first is a baseline architecture that integrates the
core components required for deployment across different hospital
environments. The second is a scalable architecture that extends the baseline
with orchestration mechanisms to support high workloads through dynamic
replication. Both versions share the same functional modules but differ in
how resources are managed and distributed under concurrent access.

\subsubsection{Baseline Architecture}

\begin{figure}[h]
    \centering
    \includegraphics[width=1\linewidth]{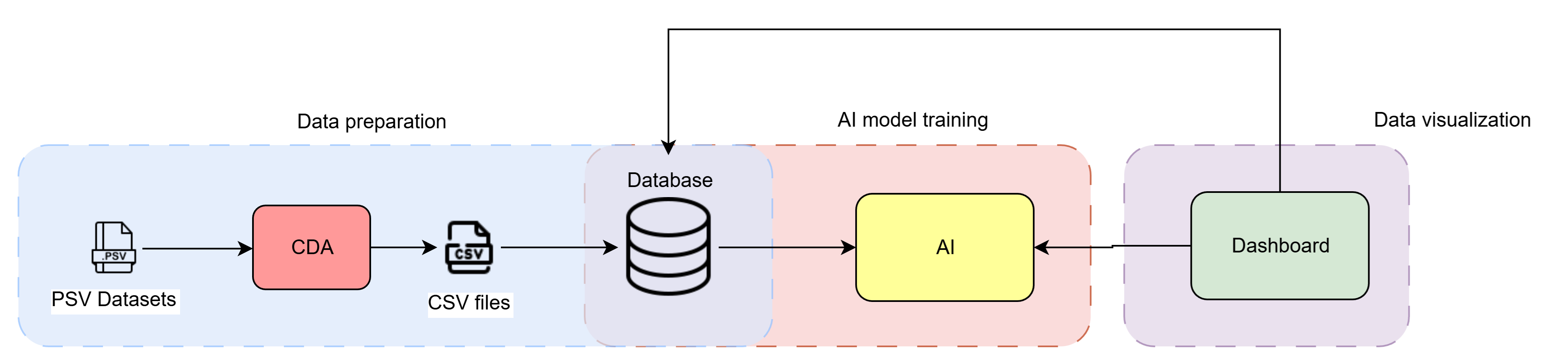}
    \caption{Baseline architecture of the SepsisAI-Orchestrator, integrating
    CDA preprocessing, MongoDB, AI service, and dashboard. Each component
    runs in an independent Docker container.}
    \label{fig:pipeline}
\end{figure}

The baseline version integrates four containerized modules, as shown in Fig.~\ref{fig:pipeline}. The first module is the \textbf{CDA Preprocessing
Service}, which transforms raw patient records into the HL7 FHIR-inspired format to ensure data standardization across hospitals. The second is the
\textbf{MongoDB Database}, a NoSQL store that holds the structured patient data in JSON format for efficient querying and aggregation. The third is the
\textbf{AI Prediction Service}, a containerized LightGBM model exposed through REST APIs using FastAPI, which receives patient identifiers and
returns sepsis risk scores. The fourth is the \textbf{Monitoring Dashboard}, a Streamlit-based web interface that allows clinicians to visualize patient
data and prediction results in real time.

Each module runs inside an independent Docker container. This design ensures
portability and reproducibility, allowing the platform to be deployed across
heterogeneous hospital environments without modifications to the underlying
code.

\subsubsection{Scalable Architecture with Orchestration}

\begin{figure}[h]
    \centering
    \includegraphics[width=1.0\linewidth]{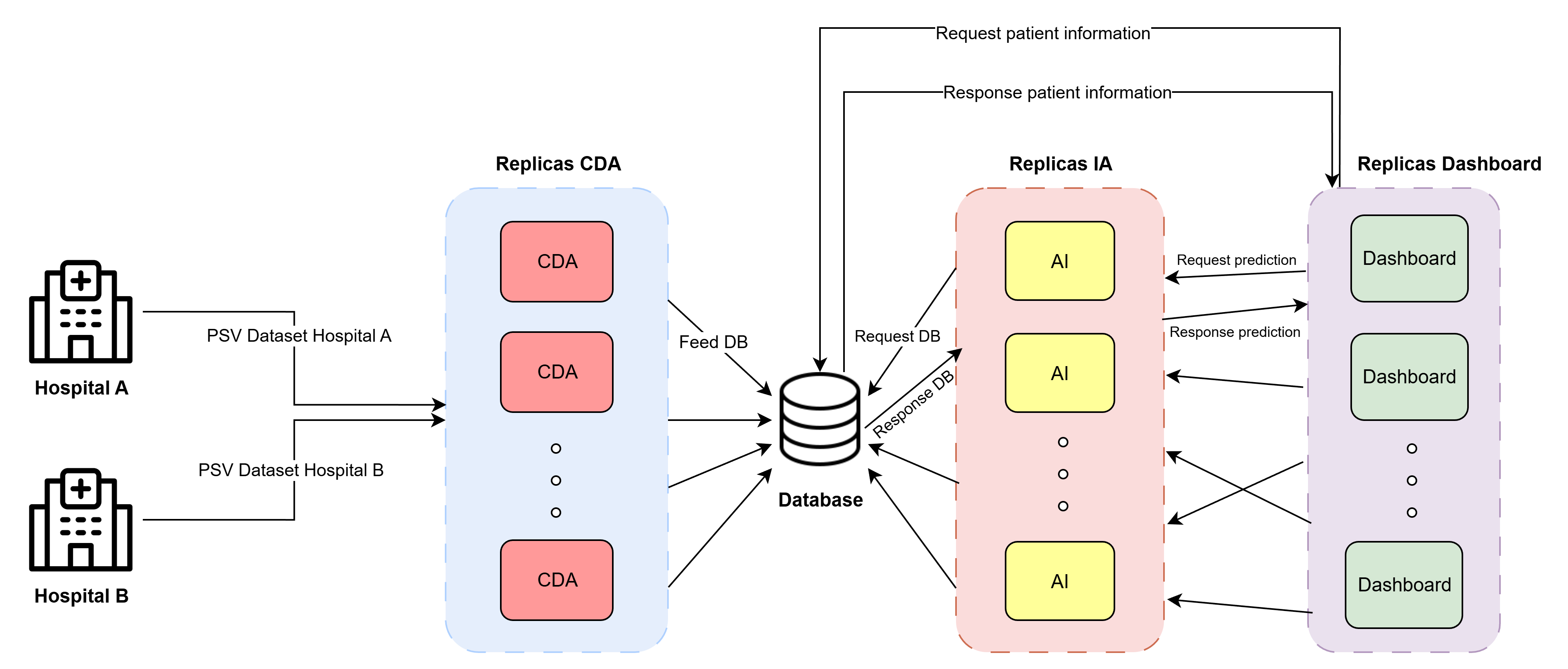}
    \caption{Kubernetes-based scalable architecture of the
    SepsisAI-Orchestrator. AI and dashboard modules are deployed as
    \textit{Deployments}, MongoDB as a \textit{StatefulSet}, CDA tasks
    as \textit{Jobs}, and \textit{Services} manage inter-component
    communication.}
    \label{fig:kubernetes_arch}
\end{figure}

The scalable version extends the baseline architecture by adding Kubernetes
orchestration for automated deployment, load balancing, and replica
management, as illustrated in Fig.~\ref{fig:kubernetes_arch}. In this
configuration, the AI Prediction Service can be scaled horizontally by
increasing the number of replicas according to available hardware resources
and user demand. The dashboard and preprocessing services can also be
replicated to handle higher concurrent loads.

Each component type is mapped to an appropriate Kubernetes resource.
The AI service and dashboard are deployed as \textit{Deployments}, which
allow Kubernetes to maintain the desired number of running replicas and
restart failed instances automatically. MongoDB is managed as a
\textit{StatefulSet} to preserve data consistency and stable network
identities across pod restarts. The CDA preprocessing tasks are executed
as \textit{Jobs}, since they run once per dataset and do not require
persistent availability. \textit{Services} provide internal communication
between all components through stable endpoints, regardless of how many
replicas are active.

This architecture supports thousands of concurrent users and provides the
fault tolerance required for large-scale hospital network deployments. The
separation between stateless compute components (AI service, dashboard) and
stateful storage (MongoDB) enables independent scaling of each layer based
on the observed bottleneck.

\subsection{SepsisAI-Orchestrator Workflow}

The end-to-end workflow of the SepsisAI-Orchestrator consists of three
sequential stages: data ingestion and standardization, AI inference, and
clinical visualization. Fig.~\ref{fig:pipeline} illustrates the data flow
across these stages.

In the first stage, patient records in pipe-separated value (PSV) format are
ingested by the \textbf{CDA Preprocessing Service}. This service transforms
the raw records into HL7 FHIR-inspired Clinical Document Architecture (CDA)
structures and stores them in the \textbf{MongoDB Database}.

In the second stage, a clinician selects a patient through the
\textbf{Monitoring Dashboard}, which sends a prediction request to the
\textbf{AI Prediction Service}. The service retrieves the corresponding CDA
record from MongoDB, extracts the relevant clinical features, and computes a
sepsis risk score using the trained LightGBM model. The predicted probability
and any triggered alerts are returned to the dashboard.

In the third stage, the dashboard displays the prediction results alongside
the patient's clinical history, enabling the clinician to assess the sepsis
risk in context.

This modular workflow decouples data standardization, model inference, and
visualization into independent services. As a result, each component can be
developed, tested, and scaled separately, which improves maintainability and
facilitates deployment across different hospital environments.

\subsection{CDA Preprocessing Service}

The CDA Preprocessing Service enables interoperability across heterogeneous
hospital systems by transforming raw electronic health records (EHR) into a
unified format. Records from hospitals A and B were converted into HL7
FHIR-compliant Clinical Document Architecture (CDA) documents through a
pipeline that performs three main operations.

First, the pipeline verifies each record for completeness and consistency.
Second, it applies temporal alignment to synchronize measurements taken at
different time points. Third, it groups clinical variables into three
categories: vital signs, laboratory results, and demographic information.
Missing values were handled using interpolation and statistical aggregation
methods described in~\cite{toro2022mlfhir}. Derived clinical scores,
including SOFA and SIRS, were computed and appended to each patient record.

The resulting standardized resources were serialized into JSON format and
stored in the MongoDB Database. This structured representation enables
efficient querying and aggregation by downstream services, particularly the
AI Prediction Service.

\subsection{AI Prediction Service}

The AI Prediction Service performs real-time inference and serves as the
connection between the data layer and the clinical dashboard. It is
implemented as a containerized LightGBM model wrapped in a FastAPI
application that exposes REST endpoints for prediction requests.

When the service receives a request containing a patient identifier, it
queries MongoDB for the corresponding CDA record. It then extracts the
required clinical features and passes them to the LightGBM model, which
computes the sepsis risk score. The result is returned to the requesting
client as a JSON response containing the predicted probability and any
associated alerts.

In addition to serving real-time requests from the dashboard, the service
can operate independently through a command-line interface (CLI). This mode
supports batch inference, model validation, and standalone evaluation
without requiring the full platform to be running. The service is packaged
in a Docker container and managed by Kubernetes, which enables horizontal
scaling through replica creation and automatic load balancing across
instances.

\subsection{Monitoring Dashboard}

The Monitoring Dashboard is the visualization layer of the
SepsisAI-Orchestrator and serves as the primary interface for clinical
users. It is built with Streamlit and integrates two data flows. The first
flow retrieves patient records from MongoDB in HL7 FHIR-inspired format.
The second flow obtains sepsis predictions from the AI Prediction Service
through its REST API.

The interface refreshes automatically every 10 seconds to reflect the most
recent data. It displays patient demographics, vital signs, laboratory
results, derived clinical scores (SOFA, SIRS), temporal timelines, and
sepsis probability estimates. When a prediction exceeds a predefined risk
threshold, the dashboard generates a visual alert to notify the clinician.

\begin{figure}[htbp]
    \centering
    \includegraphics[width=1\textwidth]{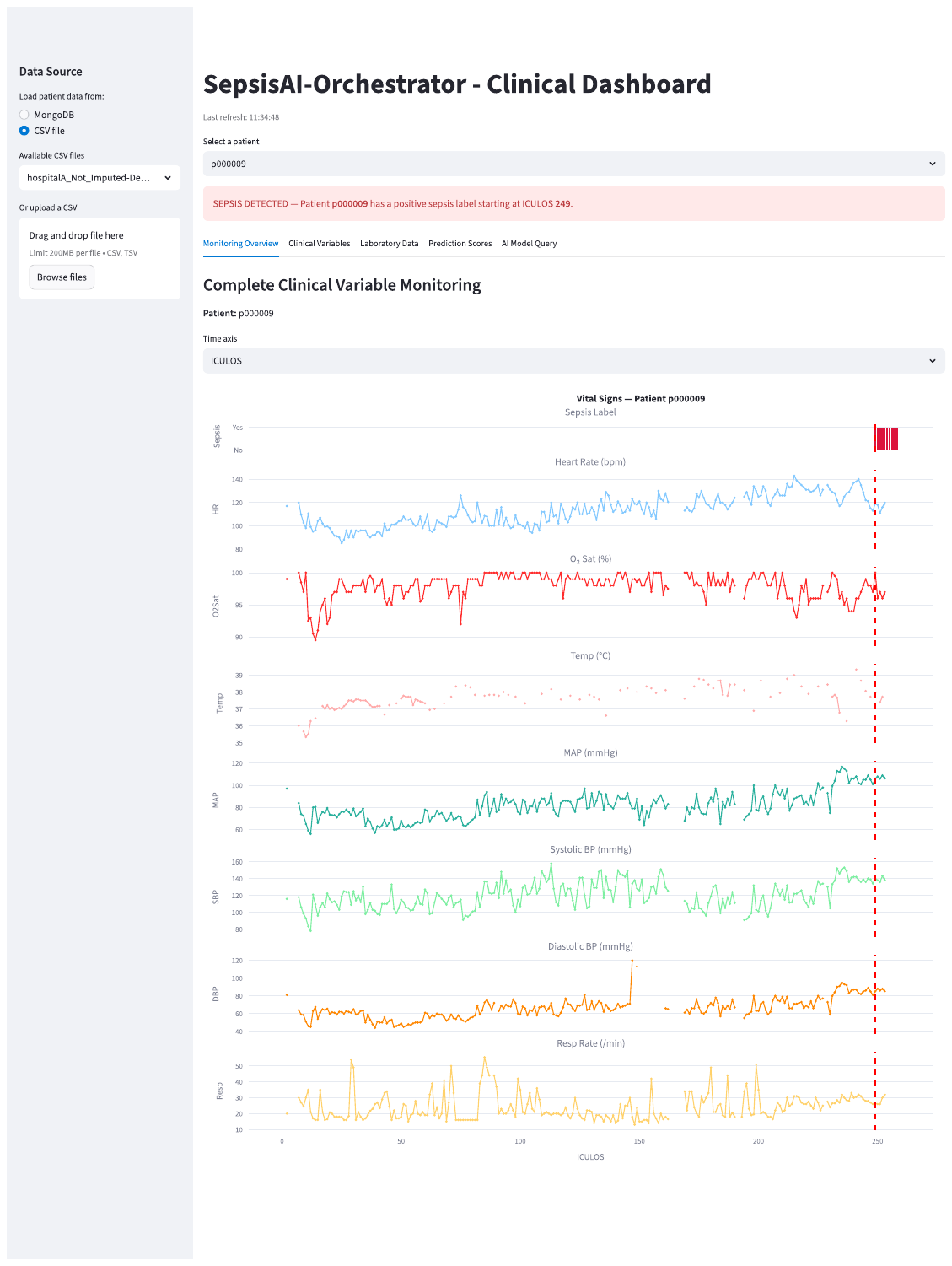}
    \caption{Monitoring Dashboard of the SepsisAI-Orchestrator showing
    patient p000009. The interface displays a sepsis detection alert
    (top banner), followed by time-series plots of eight vital signs
    (heart rate, O\textsubscript{2} saturation, temperature, mean
    arterial pressure, systolic and diastolic blood pressure, and
    respiratory rate) over ICU length of stay (ICULOS). The left panel
    allows data source selection (MongoDB or CSV), and navigation tabs
    provide access to clinical variables, laboratory data, prediction
    scores, and AI model queries.}
    \label{fig:dashboard01}
\end{figure}

The dashboard runs inside a Docker container managed by Kubernetes, which
ensures scalability and high availability during concurrent use by multiple
clinicians. Fig.~\ref{fig:dashboard01} shows the clinical dashboard in
operation.


\subsection{Performance and Scalability Analysis}
The performance of the SepsisAI-Orchestrator was evaluated using load testing with \texttt{k6}, an open-source tool for simulating concurrent users and generating reproducible stress-test scenarios. The analysis focused on scalability and responsiveness under increasing workloads, using two main metrics: throughput (requests per second) and end-to-end latency (p95 response time). 
In this study, end-to-end latency is defined as the complete execution path from a REST prediction request issued by the dashboard to the delivery of the corresponding response after data retrieval, model inference, and result visualization.
Additional indicators included failure rate (\%), iteration execution time, and network traffic to further characterize system behavior under load.
Scalability experiments simulated 50, 100, and 1000 virtual users (VUs), representing clinicians issuing real-time prediction requests. To assess Kubernetes orchestration, the AI inference service was deployed with varying replicas, and system logs were analyzed to confirm balanced load distribution. These experiments allowed both baseline validation of functionality and extended evaluation of system robustness under high concurrency.

\section{Experiments and Results}
\label{sec:experiments}

The SepsisAI-Orchestrator was evaluated in terms of interoperability, baseline functionality, and scalability under increasing workloads. The evaluation combined data preprocessing and CDA standardization, AI inference validation, load testing with k6, Kubernetes orchestration analysis, and usability validation through the monitoring dashboard.


\subsection{Experimental Setup}
Experiments were conducted on a workstation equipped with an AMD Ryzen 5 5600 CPU, 32\,GB RAM, NVIDIA RTX 4060 GPU, 1\,TB SSD, and Windows 11. The SepsisAI-Orchestrator was deployed on a Kubernetes cluster, where all components (CDA preprocessing service, MongoDB database, AI prediction service, and monitoring dashboard) were encapsulated in Docker containers. \\

The ML approach builds on previous work on missing data imputation and FHIR interoperability for sepsis prediction \cite{toro2022mlfhir}. That study compared Gradient Boosting Classifier (GBC) and Light Gradient Boosting Machine (LightGBM) models, with and without kNN-based imputation. LightGBM with imputation consistently outperformed the other methods. For instance, when trained on hospital A, LightGBM achieved an F1-score of 0.87 and AUC of 0.91, compared to 0.77 and 0.83 with GBC. Similarly, when trained on hospital B, LightGBM reached an F1-score of 0.88 and AUC of 0.91, surpassing GBC’s 0.76 and 0.82. The best results were obtained by training on the combined data from both hospitals, where LightGBM achieved F1-scores of 0.93 and 0.94, with corresponding AUC values of 0.95. \\ 

Based on these outcomes, LightGBM was selected as the prediction model and integrated into the SepsisAI-Orchestrator for deployment and scalability testing. The model was implemented as a LightGBM classifier with parameters \texttt{loss=log\_loss}, \texttt{learning\_rate=0.01}, \texttt{n\_estimators=200}, and \texttt{max\_depth=3}.

\subsection{Deployment of the SepsisAI-Orchestrator}
To assess interoperability and functionality, all clinical records from the two available hospitals in the PhysioNet dataset were processed through the CDA preprocessing service and transformed into HL7 FHIR-compliant structures. The AI prediction service was then executed on these standardized records to generate sepsis risk scores. The monitoring dashboard was used to visualize patients, clinical variables, and predicted outcomes, providing a qualitative check of the end-to-end workflow. This baseline evaluation confirmed that the SepsisAI-Orchestrator correctly ingested raw data, produced predictions, and displayed results in the dashboard.  


\subsection{Scalability Evaluation of the SepsisAI-Orchestrator}
Scalability was tested with \texttt{k6} by simulating concurrent clinicians (virtual users, VUs) issuing prediction requests. Success criteria were defined as: (i) 95\% of requests with latency $<$ 500 ms, and (ii) failure rate $<$ 1\%. Three scenarios were tested with 50, 100, and 1000 VUs. An additional experiment evaluated Kubernetes load balancing with three replicas of the AI prediction service exposed via \texttt{NodePort}. Metrics collected included throughput (requests/s), average prediction time, p95 latency, and active AI pods.\\

\subsubsection{Baseline Scalability Analysis}
The SepsisAI-Orchestrator satisfied the success criteria in the 50 and 100 VU scenarios. For example, with 50 VUs and three AI replicas, the average response time was 28.64 ms with no failures and a p95 latency of \SI{89.3}{ms}. Logs confirmed that requests were evenly distributed across replicas.  




\subsubsection{High-Concurrency Scalability Analysis}
At higher concurrency, the system showed performance degradation. In the 50 VU scenario, the average response time was 28.64 ms with a p95 latency of 89.3 ms and no failures, confirming that requests were evenly balanced across replicas. At 100 VUs, performance remained within acceptable limits, with a p95 latency of 259.3 ms and a very low failure rate of 0.03\%, caused only by minor execution delays. At 1000 VUs, however, the system exceeded the latency threshold, with a p95 latency of 3.3 s and a failure rate of 17.4\%. Despite this, all responses were valid, indicating robustness of the service under stress. Iteration-level analysis showed longer execution times as concurrency increased, while network metrics reflected proportional growth in data exchange. Overall, the platform remained functional but would require optimization to support very high user loads.

\begin{figure*}[!t]
\centering

\subfloat[HTTP response time distribution. The \SI{500}{ms} p95 target (dashed) is met at 50 and 100~VUs but exceeded at 1000~VUs.\label{fig:ext-latency}]{%
\includegraphics[width=0.42\linewidth]{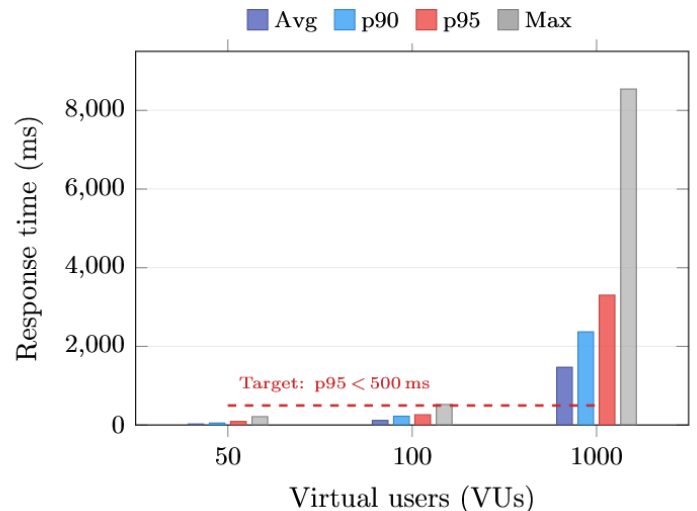}}
\hfill
\subfloat[p95 latency and failure rate vs.\ concurrency (dual axes). At 1000~VUs, p95 reaches \SI{3.3}{s} with \SI{17.4}{\percent} failures.\label{fig:ext-failure}]{%
\includegraphics[width=0.48\linewidth]{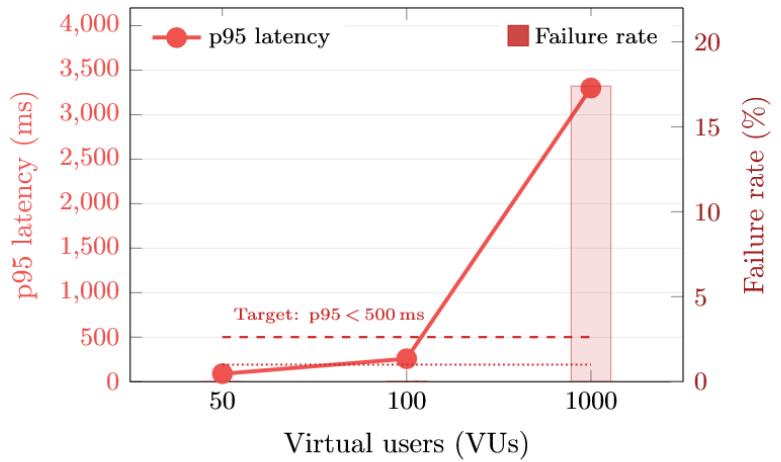}}

\vfill

\subfloat[Iteration execution metrics showing superlinear growth at high concurrency.\label{fig:ext-iteration}]{%
\includegraphics[width=0.48\linewidth]{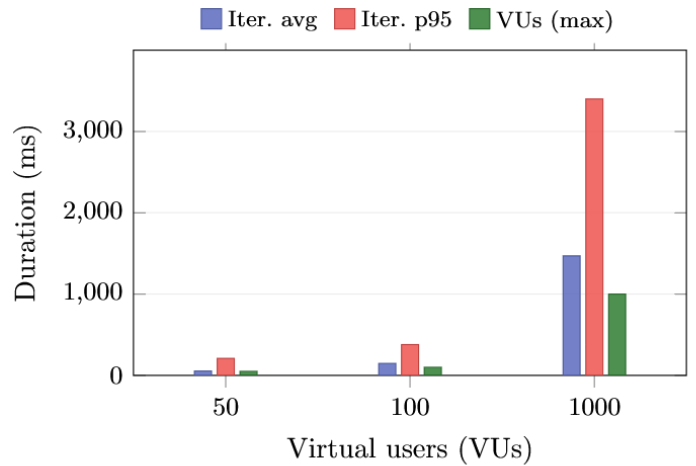}}
\hfill
\subfloat[Network traffic showing sublinear growth with concurrency.\label{fig:ext-network}]{%
\includegraphics[width=0.45\linewidth]{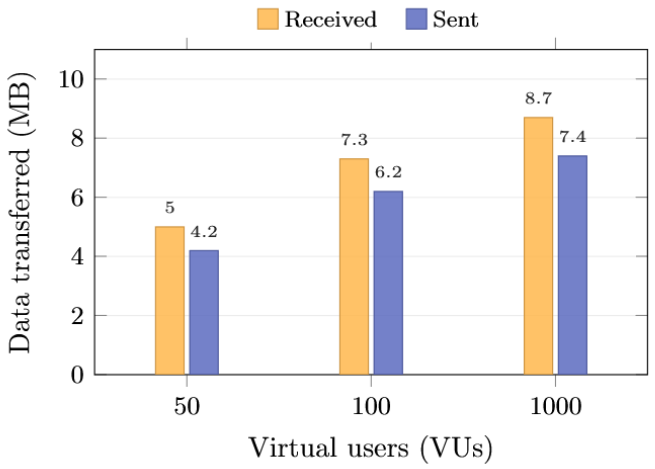}}

\vspace{4pt}

\subfloat[Consolidated baseline scalability metrics (3~AI replicas). Shaded row exceeds both success criteria.\label{fig:ext-table}]{%
\centering
\small
\setlength{\tabcolsep}{5.5pt}
\begin{tabular}{l
    S[table-format=4.2]
    S[table-format=4.1]
    S[table-format=4.1]
    S[table-format=4.0]
    S[table-format=2.2]
    S[table-format=1.1]
    S[table-format=1.1]}
\toprule
\textbf{VUs} &
\textbf{Avg (ms)} &
\textbf{p90 (ms)} &
\textbf{p95 (ms)} &
\textbf{Max (ms)} &
\textbf{Failed (\%)} &
{\textbf{Recv (MB)}} &
{\textbf{Sent (MB)}} \\
\midrule
50   & 28.64   & 52.1   & 89.3   & 210   & 0.00  & 5.0 & 4.2 \\
100  & 118.52  & 225.6  & 259.3  & 520   & 0.03  & 7.3 & 6.2 \\
1000 & 1471.29 & 2370.0 & 3300.0 & 8540  & 17.40 & 8.7 & 7.4 \\
\bottomrule
\multicolumn{8}{l}{\scriptsize Success criteria: p95\,$<$\,\SI{500}{ms} and failure rate\,$<$\,\SI{1}{\percent}. Three AI replicas via \texttt{NodePort}.} \\
\end{tabular}%
}%

\caption{Extended scalability evaluation of the SepsisAI-Orchestrator under 50, 100, and 1000 concurrent virtual users with 3~AI replicas. (a)~HTTP response time distribution across percentiles. (b)~p95 latency and failure rate on dual axes, showing both success criteria are met at 50 and 100~VUs but exceeded at 1000~VUs. (c)~Iteration execution times. (d)~Network traffic. (e)~Consolidated numerical summary. The 1000~VU results motivate the replica scaling experiments in Section~4.3.3.}
\label{fig:extended_scalability}
\end{figure*}

Fig. \ref{fig:extended_scalability} shows the extended evaluation metrics, including latency thresholds, HTTP performance, iteration execution, and network traffic. 

\subsubsection{Scalability with 1000 VUs via AI Replica Scaling}
\label{sec:replica-scaling}

The baseline evaluation with three AI replicas under 1000~VUs yielded a p95 latency of \SI{3.3}{s} and a failure rate of \SI{17.4}{\percent}, both well above the target thresholds (p95\,$<$\,\SI{500}{ms}, failures\,$<$\,\SI{1}{\percent}). To investigate whether horizontal scaling could recover acceptable performance, subsequent experiments held the load constant at 1000~VUs while varying the number of AI model replicas across 8, 12, 24, and 48 instances.

\begin{figure*}[!htb]
    \centering
    \subfloat[Response time distribution by replica count.]{\includegraphics[width=0.48\linewidth]{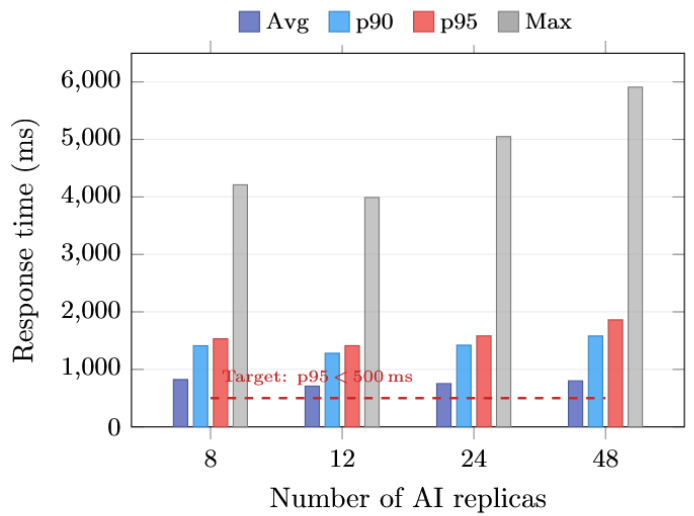}\label{fig:replica-bar}}
    \hfill
    \subfloat[p95 and average latency vs.\ replica count.]{\includegraphics[width=0.48\linewidth]{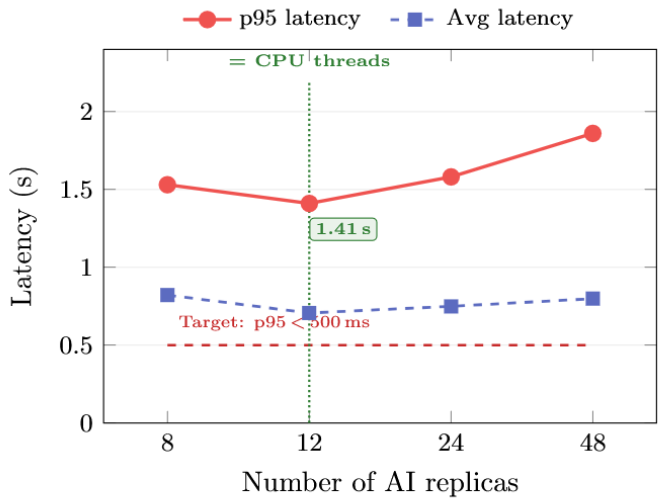}\label{fig:replica-line}}

    \vspace{6pt}

    \subfloat[Summary of HTTP performance metrics across replica configurations (1000~VUs).]{%
    \centering
    \small
    \setlength{\tabcolsep}{7pt}
    \begin{tabular}{lrrrrrl}
    \toprule
    \textbf{Replicas} &
    \textbf{Avg (ms)} &
    \textbf{p90 (s)} &
    \textbf{p95 (s)} &
    \textbf{Max (s)} &
    \textbf{Failed (\%)} &
    \textbf{$\Delta$\,p95 vs.\ 3\,rep.} \\
    \midrule
    3 (baseline)$^\dagger$ & ---   & ---  & 3.30 & ---  & 17.40 & --- \\
    8                      & 820.93 & 1.41 & 1.53 & 4.21 & 0.00  & $-53.6$\,\% \\
    12\,$\star$            & 706.38 & 1.28 & 1.41 & 3.99 & 0.00  & $\mathbf{-57.3}$\,\% \\
    24                     & 749.28 & 1.42 & 1.58 & 5.05 & 0.00  & $-52.1$\,\% \\
    48                     & 798.19 & 1.58 & 1.86 & 5.91 & 0.00  & $-43.6$\,\% \\
    \bottomrule
    \multicolumn{7}{l}{\scriptsize $\star$\,Optimal configuration (matches CPU thread count). $\dagger$\,Baseline from Section 4.3.2; only p95 and failure rate reported.} \\
    \end{tabular}
    \label{fig:replica-table}%
    }

   \caption{Effect of AI replica scaling on HTTP performance under 1000 concurrent virtual users. (a)~Response time distribution (avg, p90, p95, max) for each replica count, with the \SI{500}{ms} p95 target shown as a dashed line. (b)~p95 and average latency as a function of replica count, showing a U-shaped curve. The 12-replica configuration, which matches the CPU thread count of the test hardware (AMD Ryzen~5 5600, 6~cores, 12~threads), yields optimal performance: p95 latency of \SI{1.41}{s}, average latency of \SI{706}{ms}, and \SI{0.00}{\percent} failure rate. Increasing replicas to 24 or 48 worsens all latency percentiles due to context-switching overhead and scheduler contention. (c)~Consolidated performance metrics across all configurations. The 12-replica optimum reduces p95 latency by \SI{57.3}{\percent} relative to the 3-replica baseline and eliminates all request failures, although meeting the strict p95\,$<$\,\SI{500}{ms} threshold would require additional hardware resources or architectural optimizations.} \label{fig:different_replicas}
\end{figure*}

As shown in Fig.~\ref{fig:different_replicas}, increasing replicas from 3 to 12 produced consistent improvements across all latency percentiles: the p95 latency decreased from \SI{3.3}{s} to \SI{1.41}{s} (\SI{57.3}{\percent} reduction), the average response time dropped to \SI{706}{ms}, and the failure rate fell to \SI{0.00}{\percent}. The 12-replica
configuration corresponds to the number of hardware threads available on the AMD Ryzen~5 5600 processor (6~cores, 12~threads), representing the point at which each replica can be scheduled on a dedicated thread without contention.

Beyond this point, adding more replicas degraded performance rather than improving it. With 24~replicas, the p95 latency rose to \SI{1.58}{s} and the maximum response time increased to \SI{5.05}{s}. At 48~replicas, four times the available thread count the p95 latency further increased to \SI{1.86}{s} with a maximum of \SI{5.91}{s}. This U-shaped behavior (Fig.~\ref{fig:replica-line}) is consistent with CPU over-subscription:  when the number of active processes exceeds the hardware parallelism, context-switching
overhead and kernel scheduler contention outweigh the benefits of additional replicas.

Although none of the tested configurations met the strict p95\,$<$\,\SI{500}{ms} threshold, the 12-replica setup eliminated all request failures and achieved the lowest latency attainable on the available hardware. Meeting the sub-second target would likely require
either vertical scaling (higher core-count processors), infrastructure-level optimizations (e.g., connection pooling, caching of frequent predictions), or distributing replicas across multiple nodes in a multi-node Kubernetes cluster.

\section{\label{sec:conclusions}Conclusions and Future Work}

This work introduced SepsisAI-Orchestrator, a modular and containerized infrastructure for early sepsis detection that integrates data preprocessing, database management, predictive AI, and real-time monitoring. 
By leveraging Docker and Kubernetes, the system achieved portability, scalability, and resilience, handling up to 1000 virtual users without critical service failures. 
Key features include HL7 FHIR-inspired CDA preprocessing, MongoDB storage, LightGBM inference via FastAPI, and a Streamlit-based clinical dashboard.

Beyond demonstrating technical feasibility, the results highlight several broader implications.
The platform shows that predictive models can be moved from research environments to near-clinical settings by addressing interoperability and scalability challenges.
Building entirely on open-source tools demonstrates that advanced healthcare infrastructures can be developed with limited resources, lowering entry barriers for hospitals with constrained budgets.
While the architecture is generic by design, sepsis was intentionally selected as a representative, high-stakes clinical use case requiring strict interoperability, low latency, and continuous monitoring. As a result, the modular design makes the platform adaptable to other clinical prediction tasks and decision-support scenarios.
While prior studies predominantly focus on predictive models, data preprocessing pipelines, or isolated clinical decision-support systems, this work provides an end-to-end, containerized platform that integrates interoperability (HL7 FHIR), scalable orchestration, real-time inference, and performance evaluation within a single open-source architecture.
Stress tests revealed that while Kubernetes scaling improves performance, hardware resource limits must be carefully matched to replica counts to avoid negative overhead. This insight is relevant for institutions planning real-world deployments.
The monitoring dashboard illustrates the importance of intuitive visualization tools for clinicians, which are crucial for adoption and trust in AI-based decision support systems.

Although this study does not involve direct patient deployment, the proposed architecture is designed in alignment with HIPAA/GDPR principles through container isolation, role separation, and controlled data access.

In summary, the SepsisAI-Orchestrator demonstrates that robust and interoperable healthcare AI infrastructures are attainable with current open-source technologies, provided that attention is given to both the computational and clinical integration dimensions.

Future work will focus on: (i) clinical validation with retrospective data in hospital environments, (ii) broader interoperability with HL7/FHIR, SNOMED-CT, and OMOP-CDM, (iii) explainability to improve clinical trust, (iv) compliance with ethical and regulatory frameworks (HIPAA, GDPR), (v) continuous model adaptation, and (vi) extension to sepsis risk detection in other departments such as emergency care. These efforts aim to foster real-world adoption and improve patient outcomes.

\section*{Acknowledgements}
The authors thank AMD Colombia and AMD for providing server infrastructure support.

\bibliographystyle{apalike}
\bibliography{references}

\end{document}